%% file: main.tex
\documentclass[sigconf,natbib=true,anonymous=false]{acmart}

\input{preamble}
\usepackage{tikz, tcolorbox}
\usepackage[normalem]{ulem}

\newcommand{\DrawLine}{%
  \begin{tikzpicture}
  \path[use as bounding box] (0,0) -- (\linewidth,0);
  \draw[color=red!75!black,dashed,dash phase=2pt]
        (0-\kvtcb@leftlower-\kvtcb@boxsep,0)--
        (\linewidth+\kvtcb@rightlower+\kvtcb@boxsep,0);
  \end{tikzpicture}%
  }

\begin{document}

\title[Science Checker]{
Science Checker: Extractive-Boolean Question Answering For Scientific Fact Checking
}

\input{author}

\begin{abstract}
    \input{sections/abstract}
\end{abstract}


\input{sections/keywords}

\maketitle

\input{sections/introduction}

\input{sections/related}

\input{sections/task_formal}

\input{sections/approaches}

\input{sections/exp_res/exp_res}

\input{sections/conclusion}

\begin{acks}
We would like to thank the Vietsch Foundation for supporting the entire project.
\end{acks}

\bibliographystyle{ACM-Reference-Format}
\bibliography{references}

\end{document}

%% file: preamble.tex
\usepackage{multirow}
\usepackage[ruled,vlined]{algorithm2e}
\usepackage{cleveref}[2012/02/15]
\crefformat{footnote}{#2\footnotemark[#1]#3}
\usepackage{dsfont}

\graphicspath{{images/}}

\AtBeginDocument{%
  \providecommand\BibTeX{{%
    \normalfont B\kern-0.5em{\scshape i\kern-0.25em b}\kern-0.8em\TeX}}}

\copyrightyear{2022}
\acmYear{2022}
\setcopyright{acmlicensed}
\acmConference[MAD '22]{Proceedings of the 1st
International Workshop on Multimedia AI against Disinformation}{June 27--30,
2022}{Newark, NJ, USA}
\acmBooktitle{Proceedings of the 1st International Workshop on Multimedia AI
against Disinformation (MAD '22), June 27--30, 2022, Newark, NJ, USA}
\acmDOI{10.1145/3512732.3533580}
\acmISBN{978-1-4503-9242-6/22/06}

%% file: author.tex
\author{Loïc Rakotoson}
\email{loic.rakotoson@opscidia.com}
\orcid{0000-0002-9420-8798}
\affiliation{%
  \institution{Opscidia}
  \city{Paris}
  \country{France}
}

\author{Charles Letaillieur}
\email{charles.letaillieur@opscidia.com}
\affiliation{%
  \institution{Opscidia}
  \city{Paris}
  \country{France}
}

\author{Sylvain Massip}
\email{sylvain.massip@opscidia.com}
\orcid{0000-0002-2084-7618}
\affiliation{%
  \institution{Opscidia}
  \city{Paris}
  \country{France}
}

\author{Fréjus A. A. Laleye}
\email{frejus.laleye@opscidia.com}
\orcid{0000-0003-0744-642X}
\affiliation{%
  \institution{Opscidia}
  \city{Paris}
  \country{France}
}

\renewcommand{\shortauthors}{Rakotoson et al.}

%% file: sections/abstract.tex
With the explosive  growth  of scientific publications, making the synthesis of scientific knowledge and fact checking becomes an increasingly complex task.

In this paper, we propose a  multi-task approach for verifying the scientific questions based on a joint reasoning from facts and evidence in research articles. We propose an intelligent combination of (1) an automatic information summarization and (2) a Boolean Question Answering which allows to generate an answer to a scientific question from only extracts obtained after summarization.

Thus on a given topic, our proposed approach conducts structured content modeling based on paper abstracts to answer a scientific question while highlighting texts from paper that discuss the topic. We based our final system on an end-to-end Extractive Question Answering (EQA) combined with a three outputs classification model to perform in-depth semantic understanding of a question to illustrate the aggregation of multiple responses. With our light and fast proposed architecture, we achieved an average error rate of 4\% and a F1-score of 95.6\%. Our results are supported via experiments with two QA models (BERT, RoBERTa) over 3 Million Open Access (OA) articles in the medical and health domains on Europe PMC.

%% file: sections/keywords.tex
\keywords{Scientific Fact checking, Question Answering, Medical information extraction, Query Analysis}

%% file: sections/introduction.tex
\section{Introduction}

For many years, public trust in science has been one of the concerns, and sometimes fears, of the scientific community \cite{Haerlin, medianarrative, gaptrust}. From simple misconception to distrust, from discredit to the creation of an alternative science, opposition to science is increasingly reinforced nowadays.

Hence, the development of Open Access should be a solution to improve the communication of scientific result to the general public. Yet, even with completely open access, two important difficulties remain: domain knowledge is needed to understand the literature, and the volume of research article is such that it is difficult to read and analyse every article on a topic.

Modern communication means and the recent evolution in science communication gave the possibility to experts to popularize knowledge, notably in public conferences
and social networks \cite{medecineyoutube, tedtalk, socialpopul}. However, this has also given a voice to non-experts and pseudo-science. Science being a slow process, this gives time for obscurantism to take hold. This results in mistrust which sometimes adds to ignorance and contributes to
the scientific fake news expansion.

Therefore, automatic means to source and verify the information is more and more needed, and can be a nice complement to the work of human expertise. 
This work is part of a project called Science Checker, which aims to help non experts to navigate in the scientific literature and improve its access to the best scientific information. The general motivation and approach of the project is described elsewhere \cite{niso,oa}. In this article, we focus on the description and analysis of the classification pipeline which aims to classify articles as supporting or contradicting a scientific affirmation.

\begin{figure}
    \centering
    \begin{tcolorbox}
Question: Does Hydroxychloroquine cure COVID-19 ?
\tcblower
Abstract 1: Background Some disease-modifying agents [...] \uline{ No significant difference was found in terms of rates of usage of hydroxychloroquine or colchicine between those who were found positive for SARS-CoV-2 and those who were found negative} (0.23\% versus 0.25\% for hydroxychloroquine, and 0.53\% versus 0.48\% for colchicine, respectively)  Conclusion \uline{ These findings raise doubts regarding the protective role of these medications in the battle against SARS-CoV-2 infection.}

Answer 1: No
\par\noindent\rule{\textwidth}{0.1pt}

Abstract 2: [...] Conclusions \uline{ Hydroxychloroquine has received worldwide attention as a potential treatment for covid-19 because of positive results from small studies.} However, the results of \uline{this study do not support its use in patients admitted to hospital with covid-19 who require oxygen.}

Answer 2: No
\end{tcolorbox}
    \caption{Example of task. Given the question, the system is required to find the candidates pieces of texts and to give an answer. (Abstract 1 \cite{abstract1}, Abstract 2 \cite{abstract2})
    }
\label{fig:example}
\end{figure}

Overall, our science checker system operates on two linked tasks: (1) an extractive question answering component that, by neural semantic matching between the representations of a query and all available abstracts of scientific articles, gives an optimal representation at the granularity of text chunks for fact retrieval; (2) a multi label classification model that, from the facts found, gives the probabilities that theses are {\bf False}, {\bf True} or {\bf Neutral} with a weighted average of the results as output. Figure \ref{fig:example} perfectly illustrates the task tackled in this work behind which one can imagine the difficulties for the accurate selection of potential texts containing scientific facts from a huge knowledge base. 

To contribute to Open Science, our results, data and code will be made publicly available under a Free/Libre Open Source license.

%% file: sections/related.tex
\section{Related works}

\paragraph{Fact checking}

The SemEval-2019 fact-checking competition \cite{semeval} in Task 8B highlighted approaches by classifying terms and authors in a forum to obtain the veracity of information in a community.
The platform built by \citet{autonewsroom} allows the fact-checking of a claim by selecting the closest sentences to it within a threshold, from about ten thousand of newspaper articles, and then classifying them if they refute or support the claim. The classifier was trained on the Wikipedia-based fact-checking dataset FEVER (\citet{fever}). Similar work in \cite{NASIR2021100007} proposed an hybrid CNN-RNN model to detect the sentence that may be the fake news. \citet{externalfactchecking}’s work starts with a sentence which is preprocessed to create a query on Google and Bing search engines. The snippets in the results are then compared to the source sentence to determine whether it is factual or not. The model which compute the comparison is a combination of SVM and a recurrent neural network, and achieves excellent performance based on a dataset built from \href{https://www.snopes.com/}{Snopes}.
The approach proposed by \citet{ticnn} aims at extracting information by combining text and image which is processed by a convolutional neural network, based on their explicit and latent variables and which performs an F1 score of 92\%.

In this work and unlike the works \cite{semeval, autonewsroom, NASIR2021100007}, we focus on verifying the facts by collecting information from the scientific literature and rendering it with an intelligible answer. In contrast to newspaper information and more oriented document, the answer to a query is very rarely explicit in scientific article. Closed questions do not get a definite binary solution in the scientific literature. So, unlike the work in \cite{scifact} which similarly worked on the same task, our approach does not aim only to report the information but also to identify the interesting elements that allow for a clear answer.

In \cite{gear}, the authors proposed a graph-based information aggregation approaches from a claim to select the relative information from Wikipedia by representing it on a connected evidence graph (GEAR) for verifying a query.  

On the same basis, \citet{kgat}'s proposed an extension of the graph-based approach (GEAR) for a finer-grained highlighting of claim responses. On FEVER dataset, they outperform GEAR approach of 3.3\%. 

We deviate, in this work, from these approaches in the sense that the interest of our task is to bring out the information by scientific document so as to allow an easy tracing of the information sources.

\paragraph{Question-answering}
In this paragraph, we cite some question answering works that have adopted an extrative approach.

\citet{cloze} and \citet{synthetic} developed an approach that consists in creating an extractive question answering system without data annotations. This consists in a first step of question generation by masking or in zero-shot \cite{zeroshotGen}, followed by a second step of question answering on the previously generated data. This method avoids the need to search the dataset for a specific sub-task but requires the existence of the data in which the models will evolve. The downside is the quality dependency between the models of each phase.
\citet{biobertbioasq} obtained the best scores during the BioASQ competition \cite{bioasq19} on question-answering task with their approach based on modifying BioBERT followed by a task specific and a post-processing layer.

Many state-of-the-art approaches have been developed to address the question answering task since the long form question answering dataset \cite{eli5} has been proposed.

These approaches and methods, unlike our work, focus exclusively on extracting information without answering the question in a straightforward and concise way for the non-specialist.
SciFact \cite{scifact} partially addresses this problem by collecting sentences that explicitly respond to a claim with an additional field that states whether or not that group of sentences supports it. However, our work aims to extract less obvious information from larger sources that are intended to be unbiased. The necessary information is found on fragmented parts all over the document and the purpose is to gather it. Finally, in addition to Yes or No responses, we want to capture the neutral conclusions frequently encountered in the scientific literature.

%% file: sections/task_formal.tex
\section{Task Formalization}

As shown in the example in figure \ref{fig:example}, for a query in a specific domain (medical and health in this work), we aim first to collect the open access scientific articles related to it.

This large corpus of documents will serve as a basis for representing the scientific consensus on this question. This representation must be detailed, motivated and easily intelligible for non-experts in the field. 

Let $\mathbb{D}$ be the set of available scientific articles, $D_i$ be  an individual abstract and $S_i^{j}$ be the $j$-th text chunk in the $i$-th. For the given fact $c_i$, and the set of abstracts $\mathbb{D} = \{D_0, D_1, ..., D_n\}$ where $D_i = \{s_i^{0}, s_i^{1}, ..., s_i^{n}\}$, the task is to provide the predicted response $(\hat{R_i}, \hat{y_i})$ where $\hat{R_i}$ represents the set of sentences of an individual abstract and $\hat{y_i} \in \{T, F, N\}$ gives the probability that the given fact is True ($T$), False ($F$) and Neutral ($N$).

%% file: sections/approaches.tex
\section{Our approach}

\subsection{Method Background }

\begin{figure*}[!htb]
  \includegraphics[width=0.9\textwidth]{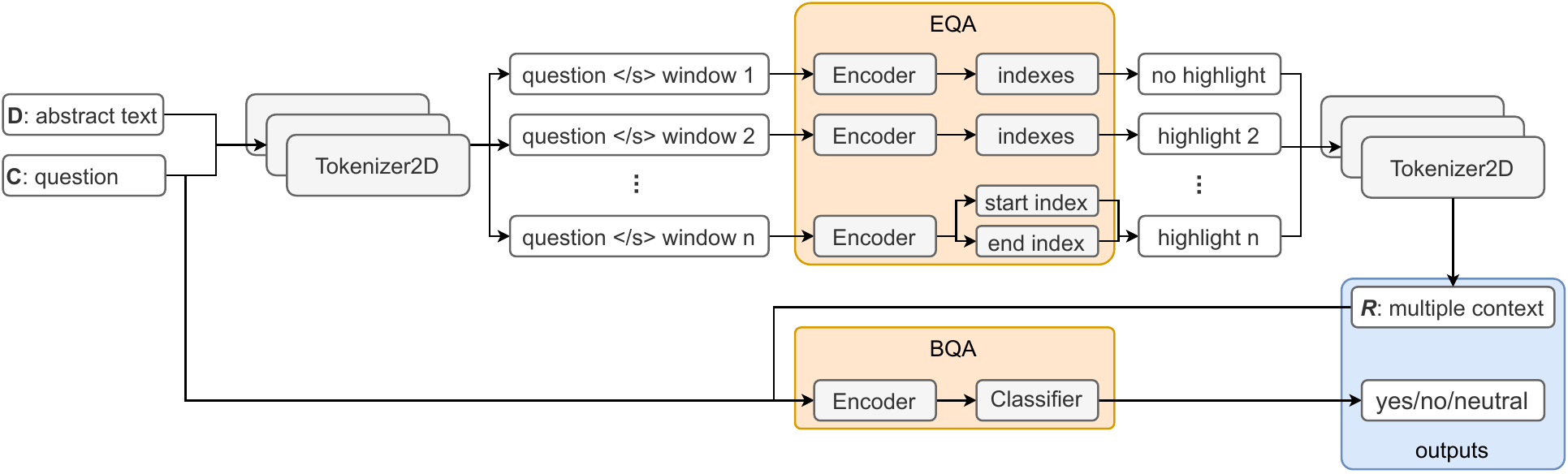}
  \caption{Combined system for science checker. \footnotesize{It consist of two two parts: abstract summarization based on Extractive Question Answering (EQA) that exploits information contained in sliding pieces of an abstract (Eq. \ref{input}) and  a multi-outputs Boolean Question-Answering (BQA) that combines all extracted contents to generate multiple output.}}
  \label{fig:pipeline}
\end{figure*}

The purpose would be to summarize multiple articles and come up with an answer (as described in the algorithm \ref{algo:algo1}). However, we want to exploit each document to be able to detail their contribution, the aggregation of which will constitute the final answer. We then propose to process each article individually, whose content can respond to the query, to extract the information needed for answering.

With the mass of open access articles on health and the thousands of possible questions, in this work we focused on closed questions worded as \textbf{\guillemotleft Does agent X \textit{prevent / cure / cause / increase} disease Y?\guillemotright}.

To retrieve the set of articles considered in our work as candidate abstracts, we perform a semantic search in the mass of open access articles using the expression for agent X and disease Y.

\begin{algorithm}
\SetAlgoLined
\KwIn{$X$: Does agent X verb disease Y?}
\KwOut{Predicted response: $(\hat{R_i}, \hat{y_i})$}
\BlankLine
Find the set of candidate abstracts $\mathbb{D}$.

\For{$D_i$ \textbf{in} $\mathbb{D}$}{
Create $n+1$ windows of $t$ sentences and stride $0 \leq p \leq t-1$ where:\\
$d = t - p$\\
$w_0 = \{s_1, ..., s_t\}$\\
$w_n = \{s_{dn+1}, ... s_{dn+t}\}$\\
where $w_j$ is truncated or padded to have 350 tokens.

    \For{$j\gets0$ \KwTo $n$}{
    $\text{EQA}_j(C, w_j) = \{s_b*, ..., s_e*\}$\\
    where $s*$ are the answers, $s_b$ the beginning of sentence highlight, $s_e$ the highlight end
    }
    
    $\hat{R_i} = \{\text{EQA}_1, \text{EQA}_j, ..., \text{EQA}_n\}$\\
	$\hat{y_i} = \text{BQA}(C, \hat{R_i})$
}
 \caption{Short summary of our proposed approach}
 \label{algo:algo1}
\end{algorithm}

The task is divided into two steps: an information gathering phase followed by a question answering phase. We used in each step Transformers-based structural textual representations \cite{attention}. The first step is to summarize an abstract based on the extraction of pieces of texts and the second step is purely a multi-outputs Boolean question-answering.

By combining all the Boolean outputs, we can therefore provide a final answer using a simple majority vote. The final output is as follows:
\begin{itemize}
    \item Affirmative: the \textit{yes} dominates.
    \item Negative: the \textit{no} prevails.
    \item Balanced: the scientific opinions are divided on the fact.
    \item Neutral: the selected OA articles do not allow to answer the question.
\end{itemize}

This combination is made to allow traceability from the response generated so that it can easily be traced back to the source articles. Figure \ref{fig:pipeline} presents an overview of our approach and shows how the two parts are linked from input representation to produce the responses. 

\subsection{Input representation}

Let $D_i$ and $N$ be respectively an abstract in the set of candidate abstracts $\mathbb{D}$ and a fixed number of word windows. We build the inputs by splitting $D_i$ into sliding windows of 350 tokens, thus forming for each abstract $D_i$, so that the $jth$ window $w_{j*}$ denotes an input feeding the models separately. $D_{i}$ is obtained as follows. 

\begin{equation}
    D_i = \sum_{j=0}^{N-1} w_{j*} 
    \label{input}
\end{equation}

\begin{figure}[!hbt]
  \centering
  \resizebox{\columnwidth}{!}{\includegraphics{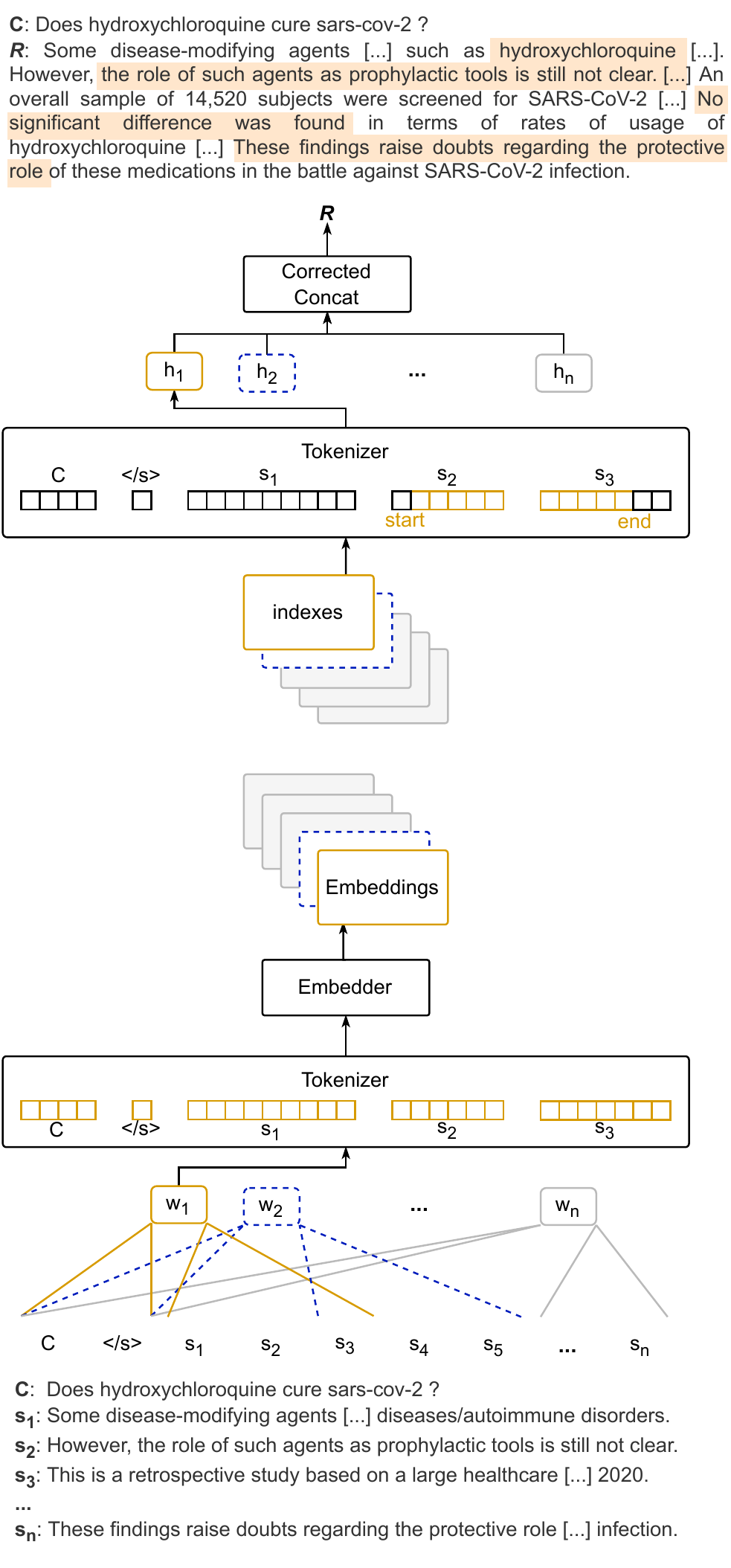}}
  \caption{
    Tokenizer2D.
    \footnotesize{The primary block (in the bottom) takes a pair of question $C$ and document $D$ decomposed into sentences $s_{1:n}$ and applies the sliding window strategy to return the embeddings. Here with a size $t=3$ and a stride $p=1$.
The secondary block (on the top) returns information to initial input $D$ from the model outputs. Here it is an extractive model indexes.}
  }
  \label{fig:tokenizer2D}
\end{figure}

Each window is fed separately to the EQA block which processes them all in parallel with as outputs: $0$ when no window allows to respond and $1$ otherwise with the spans of the resulting windows. The outputs are then concatenated (Figure \ref{fig:tokenizer2D}) to feed the BQA block. 
It should be noted that the clear answer to a query can be located in various places of the abstract (neighboring sentences or not) and that for an abstract, the answers can vary according to the input question. Our approach therefore makes it possible to address this problem by covering the information throughout the entire abstract on the one hand and by aggregating it on the other hand for a precise answer.

\subsection{Model}

In this section, we describe the core of our approach, ie the extractive and boolean models and how they have been combined.
Our information summarization model is based on an extractive approach which consists in quoting a part of the text that carries relevant information unlike the abstractive approach which consists in generating a shorter and more structured alternative text. This choice is motivated by the fact that we need the raw pieces of the text to feed the boolean model without bringing it external knowledge which could add biases to the main information and impact the decision making by the boolean model. We nevertheless experimented with the two approaches for the purposes of performance comparisons only on the information extraction.

\subsubsection{Extractive highlights}

The Extractive model consists in selecting only the relevant parts of a long text to generate a new content to be used for answering the question. Let $(C, D)$ be the set of question and text that is divided into sentences $s^*$ which are divided into tokens $\omega_{N}^* = \omega_{1}, \omega_{2}, \text{ ...}, \omega_{N}$. We then assign to each token the probabilities ($\mathbb{P}$, a matrix $\mathcal{M}$ of $N$ tokens and 2 indexes) that they are the beginning and the end of an important part. For one window, $R$ is computed such that:
\begin{equation}
\begin{split}
    &(C, D) = \omega^*_{N} \text{, } \mathds{P}\left(C, D\right) \in \mathcal{M}_{(N,\;2)} \\
    &(m, M) = \text{argmax} \left[\mathds{P}\left(C, D\right)\right] \in \mathbb{R}^2 \\
    &R = h = \omega_{m:M}
\end{split}
\end{equation}

\begin{figure}[!hbt]
  \centering
  \resizebox{.75\columnwidth}{!}{\includegraphics{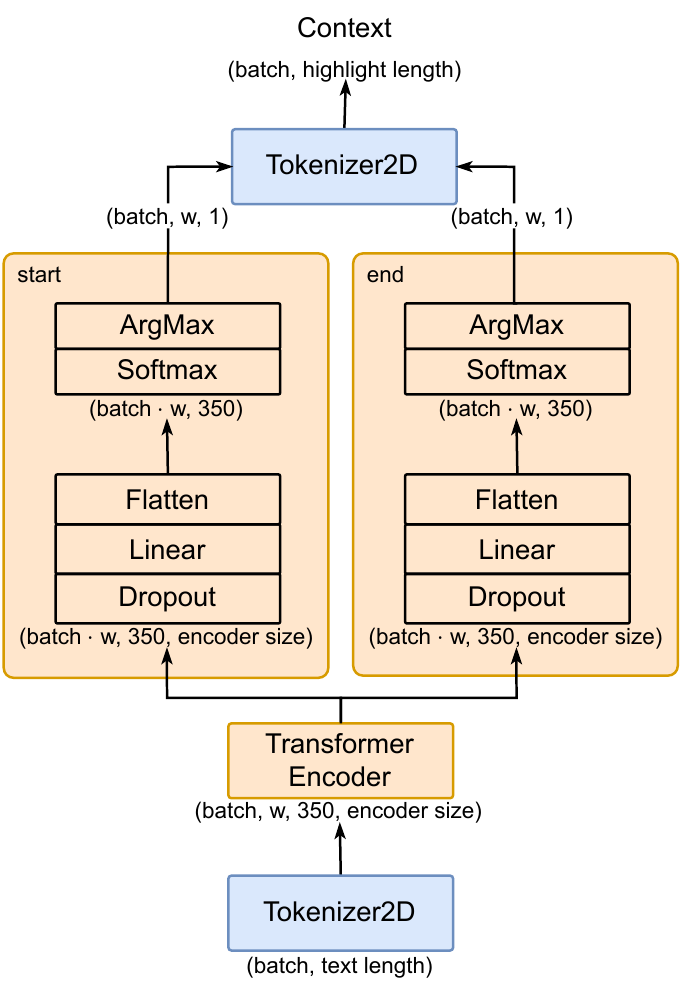}}
  \caption{
    Extractive Model.
    \footnotesize{Placed between the two blocks of Tokenizer2D, the model is composed of an Encoder followed by two outputs that assign weights to each input token.}
  }
  \label{fig:eqa_model}
\end{figure}

For each window $w_j$ of the Tokenizer2D, the couple $(C, D_i)$ gives a highlight $h_j$. Their concatenation without repetition gives the context $R$ (Figure \ref{fig:tokenizer2D}). To achieve such solution, we built a model which, starting from a text representation, outputs the start and end positions of an important part of it. We used BERT and RoBERTa \cite{roberta}, each with their base and large versions, to calculate the text embeddings. The positions are given by two dense layers of the same number of units as the input layer dimension (Figure \ref{fig:eqa_model}).

Afterwards, the best of the 4 architectures with a richer representation, and by nature more efficient than its basic version, is distilled to 2 simpler versions, MobileBERT proposed by \citet{mobilebert} and TinyBERT proposed by \citet{tinybert}. We experimented with our version of TinyBERT by keeping the same number of Encoder layers as BERT\textsubscript{BASE} but reducing the attention heads to one pair for each block. A quantization is finally applied to the optimal model.

\subsubsection{Abstractive model}

As indicated above, we also experimented with Abstractive model for summarizing information. The idea behind the Abstractive model is to take a pair of question and long document $(C, D)$, and then produce a shorter summary $R$, a sequence of tokens $\omega$ of length $T$. We then perform a directed conditional text generation such that:
\begin{equation}
\begin{split}
    R & = \omega_1, \omega_2, \text{ ...}, \omega_T \\
    R^t & = \omega_{1:t} \text{ and } R^0 = \varnothing \\
    \mathds{P}\left(R\;|\;C,D\right) & = \prod^T_{t=1} \mathds{P}\left(\omega_t\;|\;R^{t-1},C,D\right)\\
\end{split}
\end{equation}

$(C, D_j)$ are contained in each window $w_j$ of the first block of the Tokenizer2D, and all $R_j$ are concatenated in the second block of the Tokenizer2D. The found optimal value of $T$ is $80$ which does not prevent our early stopping strategy to stop the generation before $t = 80$ when the token \texttt{EOS} is part of the most likely branch.

For this type of sequence-to-sequence model, we used the auto-regressive $T5$ configuration proposed in \cite{t5} by focusing on the text summarization task. We have fine-tuned a base version with $220$ million parameters and a lighter version with about $60$ million parameters.

\subsubsection{Boolean answer}

The Boolean model is the final block which receives the extractive model outputs $R_i$, and gives a Boolean answer for each abstract $D_i$.
We fed the classifier with the text representation using the \texttt{CLS} token; and for the latter we evaluated the performance on a direct learning with the BERT architecture in its TinyBERT version and compare it with RoBERTa\textsubscript{BASE}. The model has three outputs for {\it affirmative}, {\it negative} and {\it neutral} answers only if the context does not allow to answer the question.

%% file: sections/exp_res/exp_res.tex
\section{Experiments}

First, we present the experiments with the extractive and abstractive summarization approaches before analyzing the performance of the extractive approach that best meets our task. The goal is to reduce an abstract by keeping only the relevant information.

\subsection{Experimental settings}

\subsubsection{Datasets}

To train our extractive models, we formatted the dataset to have a final context which is the concatenation of the question, the separation token and the initial context. The outputs are replaced by the answer start and end tokens positions after the final context tokenization. 

Two datasets were constituted and formatted. The first one is SQuAD v2 \cite{squad} on which the initial training was performed, the second one is the merge of BioASQ (Factoid and Lists) \cite{bioasq19} and PubMedQA \cite{pubmedqa} in which the responses are fully included in the context.

Our Abstractive training dataset is a fusion of PubMedQA  and BioASQ  for the \textit{Factoid} task, where the question and context block are concatenated with the separation token. We used Tokenizer2D with a size of $t = 7$ sentences and stride of $p = 0$ to capture the whole text, i.e., $2$ windows per article abstract at most. During the training, the input data has at most $350$ tokens, which corresponds to an average window size.

All data was transformed to BoolQ dataset format \cite{boolq} for Boolean model. We used PubMedQA dataset with the labels {\it yes/no/maybe} and BioASQ with \textit{Yes/No} task. The training dataset was augmented with Beyond Back Translation \cite{bbt} to reduce the strong imbalance between each class, then neutral contexts were additionally generated from text confusion to add material to those already existing, such that at most $30\%$ of the data is synthetic for each class. The remaining sets did not have any augmentation.

\subsubsection{Metrics}

In text generation, semantic content can be generated with different sentence structures; thus, to evaluate the results, we compared them with the paragraph written by experts using the ROUGE metric \cite{rouge} and report the $F1$-score for ROUGE-1, ROUGE-2 and ROUGE-L.

We based our extractive model evaluation on the average EM (Exact Match) score and the macro-F1 often used for this task. We also noticed that the predictions capture the true answers well but tend to add a larger or smaller margin around them; the Recall indicates the position that the relevant content takes in the predicted part.

Finally, we report Accuracy and Macro-F1 for Boolean models.

\subsection{Results}

\input{sections/exp_res/extractive}

\input{sections/exp_res/abstractive}
\input{sections/exp_res/boolean}

%% file: sections/exp_res/extractive.tex
\subsubsection{Extractive highlights}
 Table \ref{tab:eqa1} presents the results of the highlights $h_i$ extraction from each $(C, D_i)$ pair. Once $R$ is constructed as an output of the Tokenizer2D from these, the extractive approach builds a mapping over the whole document $D$ by highlighting all the important $h_i$ information scattered all over the large text. The strength of the approach lies in capturing the indirect and implicit relationships between important terms. As a result, the subsequent Boolean model will no longer need to deal with irrelevant material to answer the question but will start directly from condensed information.

\begin{table}[!htpb]
 \caption{Results of extractive models.}
  \centering
  \resizebox{\columnwidth}{!}{
  \begin{tabular}{lccccc}
    \toprule
    & \multicolumn{3}{c}{Scores} & \multicolumn{2}{c}{Statistics}  \\
    \cmidrule(r){2-4} \cmidrule(r){5-6}
    Model                    & EM    & F1    & Recall& P   & V (w/s)\\
    \midrule
    BERT\textsubscript{B}     & 40.00 & 58.35 & 72.34 & 110 & 1.49\\
    BERT\textsubscript{L}     & 41.69 & 63.32 & 74.55 & 334 & 0.58\\
    RoBERTa\textsubscript{B}  & 39.66 & 62.98 & 71.81 & 124 & 1.42\\
    RoBERTa\textsubscript{L}  & 39.39 & 62.82 & 67.01 & 355 & 0.39\\
    \midrule
    MobileBERT & 41.46 & 60.14 & 71.42 & 25 & 2.03 \textbf{x4}*\\
    TinyBERT   & 37.40 & 58.35 & 71.02 & 6  & 6.50 \textbf{x11}*\\
    \bottomrule
  \end{tabular}
  }
  \label{tab:eqa1}
 \caption*{\footnotesize Parameters (P) in Millions. Inference speed (V) in windows per second (w/s), computed on CPU: Intel Xeon @ 2.20GHz\\
 *Speed-Up compared to BERT\textsubscript{L}
 }
\end{table}

Among the architectures trained directly on the corpus, BERT\textsubscript{LARGE} performs best and infers faster than RoBERTa for the same level of complexity.
At this stage, we only consider the encoders' performance. A direct benefit would be to take the most efficient one and lighten it to a less complex and faster version while keeping its knowledge. In favor of the inference speed,  it perform at least as well as the base even if its distillations  lower its scores.

The TinyBERT version particularly performs scores close to the base with a Recall not very different from its MobileBERT counterpart, but has the advantage of being lighter and faster.

This Recall difference is justified by the fact that the target $h_j$ are contained in the TinyBERT $\hat{h}_j$ but irrelevant tokens are added before and/or after. This performance loss is considered negligible as long as the additions are minor.

%% file: sections/exp_res/abstractive.tex
\subsubsection{Abstractive summarization}

We generated context summarization for each version of the model by playing with the Beam search’s \cite{beam_s} depth and the n-grams size without repetition.

\begin{table}[!hpbt]
 \caption{Results of Abstractive model.}
  \centering
  \resizebox{\columnwidth}{!}{
  \begin{tabular}{lccccccc}
    \toprule
    Model & B & NR & \multicolumn{3}{c}{ROUGE} & \multicolumn{2}{c}{Statistics}  \\
    \cmidrule(r){4-6} \cmidrule(r){7-8}
              &    &      & 1    & 2    & L   & P & V(w/s)  \\
    \midrule
    Seq2Seq*                                & 5 & 3 & 28.9 &  5.4& 23.1 & - & - \\
    \hline
    \multirow{3}{*}{T5\textsubscript{SMALL}}& 3 & 2 & 31.3 & 13.8& 25.6 & \multirow{3}{*}{60} & \multirow{3}{*}{5.67}\\
                                        & 5 & 4 & 31.4 & 14.4& 25.9 & &    \\
                                        & 10& 4 & 31.5 & 14.3& 25.8 & &    \\
    \hline
    \multirow{2}{*}{T5\textsubscript{BASE}}& 3 & 2 & 30.4 & 13.3& 24.9 &\multirow{2}{*}{220} &\multirow{2}{*}{3.60}\\
                                        & 5 & 4 & 31.5 & 14.3& 26.0 &      \\
    \bottomrule
  \end{tabular}
  }
  \label{tab:aqa1}
 \caption*{\footnotesize Beam search (B). No-Repeat ngram size (NR). Parameters (P)  in Millions. Generation speed (V) in windows per second (w/s),
 computed with GPU: NVIDIA GeForce GTX 1050\\
 *Seq2Seq Multi-Task: performance on ELI5
 }
\end{table}

The T5\textsubscript{SMALL} version is understandably faster and, despite its lower complexity, its performance (Table \ref{tab:aqa1}) is no different from the basic version especially if we observe ROUGE-L.

In the generation, we want to avoid the insertion of insights that are not originally in the document $D$ in order not to bias the interpretation of the Boolean model. In our case, the more words the summary $R$ quotes from the article, the better. Nevertheless, this better output form is equivalent to performing extractive.

Compared to an Extractive output, the Abstractive texts are structured and fully intelligible to human.
However, the downside is the loss of raw information to trace back to the original text, which is still an important point in our approach.

%% file: sections/exp_res/boolean.tex
\subsubsection{Boolean Model and Combined System}

To fully meet our goal and based on the results of the two previous models, we plugged the Boolean model to the optimal extractive model (TinyBERT). It takes as input the concatenation of the question and the set of extracted windows outputs.

BERT and RoBERTa base version are used as encoder and we also experimented with the previously developed TinyBERT trained this time without distillation but directly on the raw data.
We additionally reported the results (Table \ref{tab:bqa1}) by first running the text through one of the summary models before the best Boolean model. These are BERT\textsubscript{L} for extractive and T5\textsubscript{S} for abstractive.

\begin{table}[!htpb]
 \caption{Results of Boolean model.}
  \centering
  \begin{tabular}{lcccc}
    \toprule
    & \multicolumn{2}{c}{Scores} & \multicolumn{2}{c}{Statistics}  \\
    \cmidrule(r){2-3} \cmidrule(r){4-5}
    Model                   & Accuracy & F1 & P   & V (a/s)\\
    \midrule
    BERT\textsubscript{B}    & 96.85 & 96.01 & 110 & 3.45\\
    RoBERTa\textsubscript{B} & 97.32 & 97.07 & 125 & 3.61\\
    TinyBERT                 & 96.55 & 95.60 & 6   & 6.09\\
    \midrule
    Ext.QA + TinyBERT        & 95.07 & 86.15 & 249 & - \\
    Abs.QA + TinyBERT        & 98.79 & 97.11 & 345 & - \\
    \bottomrule
  \end{tabular}
  \label{tab:bqa1}
 \caption*{\footnotesize Parameters (P) in Millions. Inference speed (V) in \textbf{article per seconde} (a/s), computed on CPU: Intel Xeon @ 2.20GHz
 }
\end{table}

The results in the first part of Table \ref{tab:bqa1} show the performance of using the multiple encoders based on Roberta ($F1=97.07\%$) compared to Bert. Roberta being pre-trained to better encode the distributional representation of a sentence, it therefore reinforces the performance of the boolean question answering task. The direct training with the simple text representation of TinyBERT gives quite good scores considering its gap with our best model. We chose to quantize this model directly to speed it up to 9 items per second with a minor $F1$-score degradation of a 0.27\% difference.

The second part of Table \ref{tab:bqa1} presents the performance of the combined system: the boolean model plugged into the model of information summarization. We observe that abstractive model gives the best $F1$ score ($97.11$) compared to extractive model ($86.15$). 

This performance gap reveals a specialization effect in the abstractive model that allows the boolean model to learn from richer summary representations by bringing it more consolidated knowledge. On the other hand, the extract modeling does not consolidate knowledge but produce raw extracts which are consumed by the boolean model. Based on the results in Table \ref{tab:bqa1}, the contraction of information, sometimes with noise, performed by the abstractive leads the boolean model to better answer the question, in contrast to the extractive which reduces efficiency with respect to the unsummarized inputs. However, there is a trade-off between the performance, the traceability of the gathered information and the requirements of the task formalism. 

Based on all this, the extractive model remains the best approach for our task and its combination with the boolean model provides a multi-task system that efficiently produces both:

\begin{itemize}
    \item a boolean answer to a question;
    \item the extracts from the texts (not like a black box) supporting an answer;
\end{itemize}

These performances were not compared to related works as our task differs slightly from the usual fact checking task as indicated above.

%% file: sections/conclusion.tex
\section{Conclusion}

By combining an extraction and answering system, we are able to find the solution to a closed question while justifying it from the source excerpts. The extractive part combined with our way of consolidating information through the produced extracts  with Tokenizer2D
allows to translate non-explicit information into a more concrete knowledge set. It then allowed both to give an answer to a question and to identify the relevant information for the answering system part. 

Our experiments on the significant reduction of millions of model parameters made it possible to reduce the complexity of the final model, even if we can observe a slight decrease in performance. Which leads to a multi-task system that embeds the two TinyBERT models with an average error rate of 4\%, which perform less than the one with a RoBERTa head at 2.68\% error rate, but clearly lighter and faster.

The final system takes a query and returns two outputs for each article: the highlights and the Boolean answer. 

We represent the scientific consensus on the question by using a simple majority vote of each article. Our approach is indeed to give a representation of the scientific consensus and not to give the absolute truth. However, this representation is limited by the availability of open access articles related to claims and by its quantitative aspect of the consensus with the aggregation.

We are currently work to improve the proposed approach by integrating the possibility for models to combine multiple and inevitably contradictory answers into a coherent and understandable response for humans.